\documentclass[letterpaper, 10 pt, conference]{ieeeconf}  

\usepackage{times}
\usepackage[dvipsnames]{xcolor}
\usepackage{multicol}
\usepackage[bookmarks=true]{hyperref}
\usepackage{graphicx}
\usepackage{algorithm}
\usepackage{amsmath}
\usepackage[normalem]{ulem}
\usepackage{subcaption}
\usepackage{amssymb}
\usepackage{booktabs} 
\usepackage{multirow} 
\usepackage{tcolorbox}
\usepackage[table]{xcolor}
\usepackage{siunitx}

\usepackage{booktabs}
\usepackage{caption}

\usepackage[normalem]{ulem}

\definecolor{clsGeneralist}{RGB}{229,245,224} 
\definecolor{clsSpecialist}{RGB}{253,233,232} 
\definecolor{clsOurs}{RGB}{232,240,255}       
\definecolor{clsPropGen}{RGB}{236,231,255}    
\definecolor{bestCell}{RGB}{255,244,204}      

\definecolor{capbg}{gray}{0.93}
\DeclareCaptionFormat{colorbox}{
  \colorbox{capbg}{\parbox{\dimexpr\linewidth-2\fboxsep\relax}{#1#2#3}}
}
\definecolor{envCol}{RGB}{255,236,214}    
\definecolor{agentCol}{RGB}{232,240,255}  
\definecolor{compCol}{RGB}{229,245,224}   
\definecolor{metricCol}{RGB}{236,231,255} 
\definecolor{spatialCol}{RGB}{253,233,232}

\newif\ifcomments
\commentstrue   

\newcommand{\tn}[1]{\ifcomments\textcolor{teal}{#1}\fi}

\IEEEoverridecommandlockouts                              

\title{\LARGE \bf
Meanings and Measurements: Multi-Agent Probabilistic Grounding for Vision-Language Navigation
}

\author{
\begin{tabular}{c}
Swagat Padhan$^{*,1}$, Lakshya Jain$^{*,1}$, Bhavya Minesh Shah$^{1}$, Omkar Patil$^{1}$, Thao Nguyen$^{2}$, Nakul Gopalan$^{1}$ \\[4pt]
$^{1}$Arizona State University \qquad $^{2}$Haverford College \\[2pt]
{\footnotesize \texttt{\{spadhan, ljain6, bshah43, opatil3, ngopalan\}@asu.edu, tnguyen3@haverford.edu}}
\end{tabular}
}

\begin{document}

\maketitle
\begin{figure}[b]
\vspace{-10pt}
\begin{tcolorbox}[
  colback=white,
  colframe=black,
  boxrule=0.3pt,
  arc=0pt,
  left=4pt, right=4pt, top=3pt, bottom=3pt,
  fontupper=\footnotesize
]
$^{*}$Equal contribution. Submitted to IROS 2026. \\This work was partly funded by TSMC and Arizona State University.
\end{tcolorbox}
\end{figure}
\thispagestyle{empty}
\pagestyle{empty}

\begin{abstract}

Robots collaborating with humans must convert natural language goals into actionable, physically grounded decisions. For example, executing a command such as “go two meters to the right of the fridge” requires grounding semantic references, spatial relations, and metric constraints within a $3$D scene. While recent vision language models (VLMs) demonstrate strong semantic grounding capabilities, they are not explicitly designed to reason about metric constraints in physically defined spaces. In this work, we empirically demonstrate that state of the art VLM-based grounding approaches struggle with complex metric-semantic language queries. To address this limitation, we propose MAPG (Multi-Agent Probabilistic Grounding), an agentic framework that decomposes language queries into structured subcomponents and queries a VLM to ground each component. MAPG then probabilistically composes these grounded outputs to produce metrically consistent, actionable decisions in $3$D space. We evaluate MAPG on the HM-EQA benchmark and show consistent performance improvements over strong baselines. Furthermore, we introduce a new benchmark, MAPG-Bench, specifically designed to evaluate metric-semantic goal grounding, addressing a gap in existing language grounding evaluations. We also present a real-world robot demonstration showing that MAPG transfers beyond simulation when a structured scene representation is available. 
\end{abstract}

\section{Introduction}

Robots that collaborate with humans must be able to convert natural language goals into actionable, grounded decisions. People mix metric and semantic specifications all the time from ``being five minutes late'' to ``taking a left after about a mile on the highway.'' 
We refer to these as metric-semantic queries. The semantics relate to the attributes of an object or concept and spatial relationships such as left, right, front, behind; whereas the metric part of the query relates to measurable quantities such as distances, time and scale.
\begin{figure}
    \centering
    \includegraphics[width=1\linewidth]{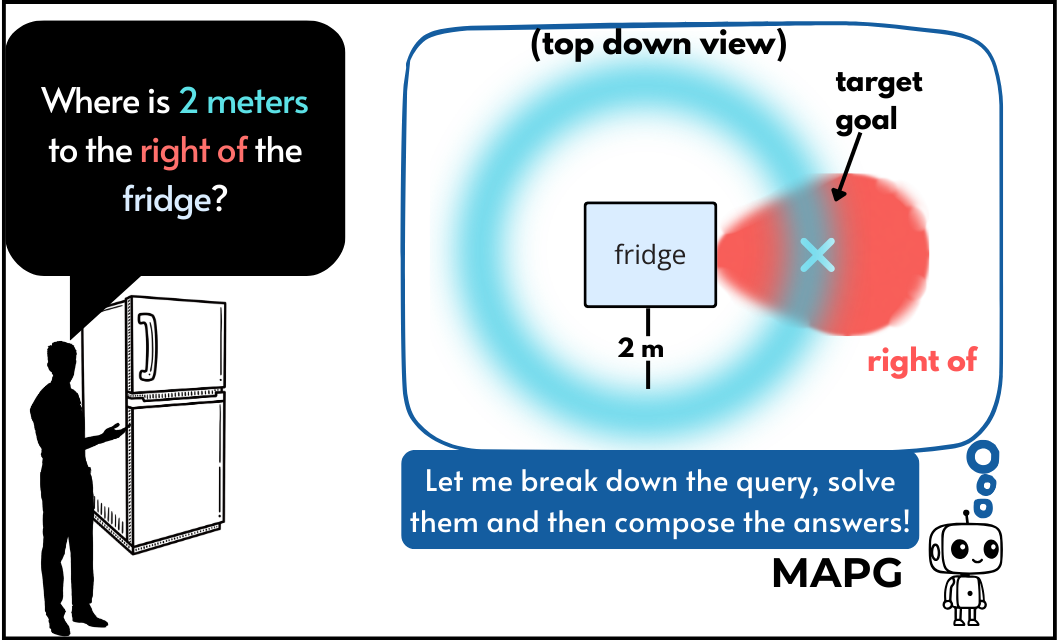}
    \caption{When a human specifies an open world metric-semantic query, referring to a location in the world, MAPG decomposes the query into subparts as learned probability distributions and composes them for grounding.
    }
    \label{fig:overview}
\end{figure}
Grounding these metric-semantic queries in a map requires jointly reasoning over referent semantics, spatial predicates, and metric scale, where the intended goal is a continuous region anchored to a grounded object pose.

Instruction following is a central subfield of grounded language learning in robotics \cite{tellexlanguagesurvey}. Most existing systems ground language to actions or goals, which can be objects in a scene \cite{dai2024thinkactaskopenworld}. 
Several grounded language systems emphasized structured decomposition and probabilistic inference to connect linguistic constituents to world entities~\cite{tellex2011symbolgrounding}  while modern navigation and embodied question answering (QA) systems increasingly rely on large vision-language models (VLMs) and large language models (LLMs) to reason over observations and internal scene representations such as $3$D semantic scene graphs \cite{graphEQA, rana2023sayplan}. 
These approaches treat goal grounding as a single-step decision: given the current observation (and sometimes an internal map), the model is prompted to output either a discrete action or a single target hypothesis \cite{vsiBench,frameOfReferenceAmbiguity}.  However, such designs can be fragile for metric-semantic instructions whose correctness depends on precise geometry and a consistent frame of reference while navigating to discover the goal.
Furthermore, language grounding is bidirectional: the agent must convert egocentric observations into allocentric positions on the map, then convert an allocentric goal back into egocentric coordinates for execution, which compounds errors from each step.

We propose MAPG (Multi-Agent Probabilistic Grounding), an agentic framework that queries different VLM Agents to produce probabilistic distributions and composes them into the goal. Given an instruction, MAPG decomposes it into structured clauses, resolves referents against an online $3$D scene graph and the current egocentric view, and instantiates analytic kernels for semantic, metric, and spatial constraints.
These kernels are then composed to produce a final goal density that better represents the spatial intention, which a planner can then query to generate executable waypoints.
Fig. \ref{fig:overview} shows an example of this approach.

Our main contributions are as follows: 
\begin{itemize}
    \item \textbf{A multi-agent probabilistic $3$D spatial reasoning framework for goal grounding} that couples online $3$D scene graphs with analytically defined continuous spatial kernels to produce planner-ready goal distributions for metric-semantic instructions.
    \item \textbf{A first-of-its-kind HM$3$D~\cite{ramakrishnan2021habitatmatterport3ddatasethm3d} goal grounding benchmark for metric-semantic queries}, including an open-source dataset and evaluation protocol spanning $30$ unique indoor scenes and $100$ annotated metric-semantic queries designed for object-to-world goal grounding in realistic indoor layouts.
    \item \textbf{Empirical findings and ablations}: We show that our method achieves low distance error (\SI{0.07}{\meter}) and low angular errors (\SI{0.3}{\degree} yaw, \SI{3.8}{\degree} pitch) in goal grounding, and additionally provide a failure taxonomy covering failure mode categories, enabling reproducible comparison for future goal-grounding systems.
\end{itemize}
Our webpage includes videos, benchmark examples, and extensive results: \href{https://sites.google.com/view/mapg-web/home}{sites.google.com/view/mapg-web/home}.
\begin{figure*}[t]
    \centering
    \vspace{1mm}
    \includegraphics[width=0.99\linewidth]{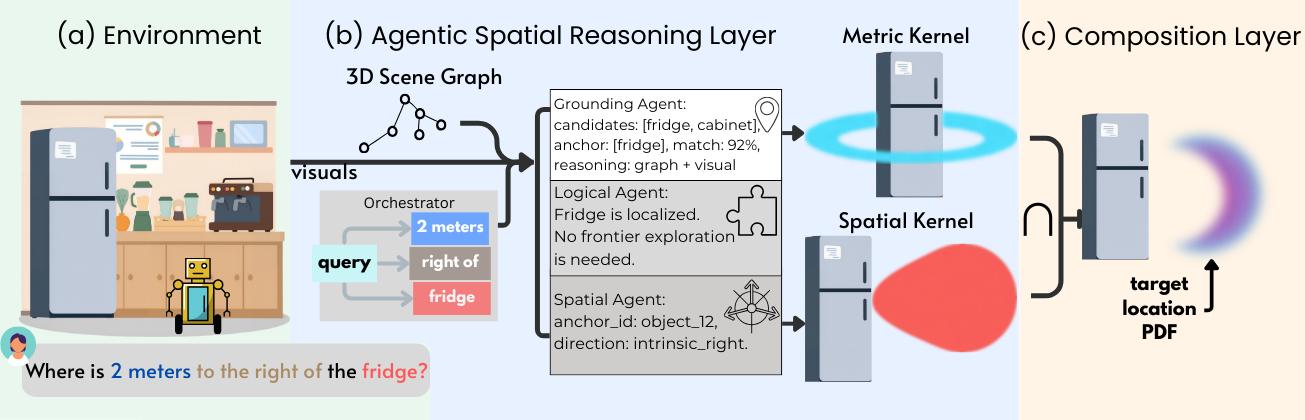}
    \caption{\textbf{MAPG overview.}
    (a) The agent observes a $3$D environment and receives a natural-language spatial query.\;
    (b) Agentic spatial reasoning layer: The Orchestrator decomposes the query into anchor, relation, and metric components and grounds the anchor instance using multi-view evidence and the $3$D scene graph.\;
    (c) MAPG composes probabilistic kernels, producing a continuous target-location PDF in the global frame that can be used as an actionable navigation or object selection goal.
    }
    \label{fig:method-overview}
\end{figure*}
\section{Background and Related Work}
Our work lies at the intersection of embodied vision-language navigation, structured $3$D spatial memory, spatial reasoning with multimodal foundation models, and probabilistic grounding. We review the most relevant directions and highlight the gaps our method addresses. 

\noindent{\textbf{{Embodied Question Answering and Vision-Language Navigation:}}}
Embodied Question Answering (EQA) and Vision-and-Language Navigation (VLN) both require an agent to ground language into actions over long horizons under partial observability, using only egocentric observations~\cite{das2018embodiedqa}.
Many embodied systems still commit to a single action or discrete target at each step, so early grounding errors compound and can derail the entire trajectory \cite{cohen2024surveyroboticlanguagegrounding,ahn2022saycan, dai2024thinkactaskopenworld}.
Some prior VLN-based works explicitly identify this issue and introduce backtracking and progress-based recovery to mitigate trajectory drift under greedy decisions~\cite{ma2019regretful}.
Conceptually, our approach addresses the same failure mode by shifting goal grounding to producing a planner-ready goal representation in a generated map rather than a one-shot action choice.

\noindent{{\textbf{Structured $3$D Spatial Memory and Scene Graphs:}}}
Embodied reasoning benefits from having an explicit spatial memory that persists across time and supports queries about objects, places, and their relations. Hydra~\cite{hydra}, along with Kimera~\cite{rosinol2020kimeraopensourcelibraryrealtime} construct and optimize an online $3$D scene graph that unifies metric layers (e.g., distance fields), object nodes, and higher-level place and room structure~\cite{hydra}.
However, scene graphs cannot produce planner-ready navigation goals as they contain only adjacency information while lacking metric information about a physical space. We therefore build on Scene Graphs a spatial backbone over which metric-semantic language can be composed into planner-ready target distributions (see Fig. \ref{fig:method-overview}).

\noindent\textbf{Spatial Reasoning Limits of Multimodal Foundation Models:}
Although LLMs and VLMs are strong on many language tasks, their spatial competence remains inconsistent~\cite{vsiBench,frameOfReferenceAmbiguity,spatialRGPT, 3dsrbench}. Across evaluations, common failure modes include unstable visual distance estimates, inconsistent camera framing, and lack of mechanisms to ground visual information \cite{frameOfReferenceAmbiguity, 3dsrbench}.
Several works attempt to address these issues through spatial fine-tuning or by incorporating depth, but the gap  remains for embodied settings where the agent must maintain a consistent allocentric representation over time~\cite{vsiBench,spatialRGPT, 3dsrbench}. Our approach addresses these limitations at the interface level by making the spatial target explicit within a metric semantic map.

\noindent\textbf{Vision-Language-Action Models and Affordance Learning:}
Vision-language-action (VLA) models learn end-to-end policies from multimodal inputs to actions, and large-scale training has shown strong transfer, with navigation-focused variants improving robustness through flexible goal conditioning on language, goal images, or goal poses \cite{openx2023rtx,hirose2025omnivla}. However, these policies are typically trained to predict actions rather than to explicitly infer a metric goal in a shared $3$D frame. Our work is novel in that it creates compositional interfaces between a planning or navigation stack, a semantic perception stack, and a mapping stack to generate actions. 

\noindent{\textbf{{Probabilistic Grounding and Multi-Expert Fusion:}}}
Grounding language in the physical world has long been treated as probabilistic inference, where linguistic structure induces latent grounding variables over referents, relations, and actions \cite{tellex2011symbolgrounding, ijcai2018p810}. This formulation is
well suited to spatial language, where predicates such as ``left of'' or ``near'' define distributions over regions of space rather than point locations. A complementary theme is decomposing a complex task into reusable parts to generalize better than monolithic policies when the combinatorics grow \cite{poco, patil2025factorizingdiffusionpoliciesobservation, du2024compositionalgenerativemodelingsingle}. To combine multiple information sources, products of experts (PoE) offer a principled mechanism that multiplies distributions and renormalizes, producing solutions that satisfy multiple constraints simultaneously \cite{hinton2002poe}. Finally, recent embodied systems increasingly combine multiple models or signals, such as mixing perception with language reasoning or fusing goal proposals with feasibility checks \cite{rana2023sayplan}.


Our work builds on these directions by addressing the missing interface between structured $3$D spatial memory and planner execution for metric-semantic queries, providing a map-aligned and controllable mechanism for translating language into navigation targets (see Fig. \ref{fig:method-overview}).
\begin{figure*}
    \centering
    \includegraphics[width=0.99\linewidth]{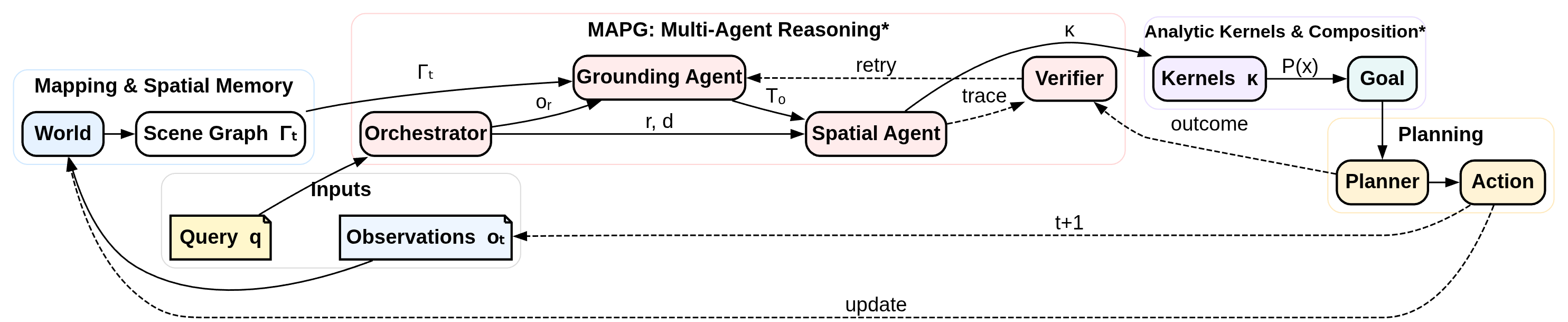}
    \caption{
      \textbf{MAPG system overview.}
      Egocentric observations $o_t$ are fused into a $3$D scene
      graph $\Gamma_t$ of labeled objects with poses and bounding boxes.
      The \emph{Orchestrator} parses a language query $q$ into
      composable symbolic predicates $(o_r, r, d)$.
      The \emph{Grounding Agent} resolves the referent $o_r$ against $\Gamma_t$,
      \emph{Spatial Agent} constructs analytic kernels
      $\kappa = \kappa_s \cdot \kappa_m \cdot \kappa_d$
      and composes them into a goal distribution $P(x)$ over free space.
      The \emph{Verifier} checks coherence and triggers corrective retries
      when needed.
      Finally, the \emph{Planner} selects a navigation action $a_t$,
      closing the loop as new observations update $\Gamma_t$. Our contributions:*  
    }
    \label{fig:method_overview-h}
\end{figure*}

\section{Methodology}
\noindent\textbf{Problem Definition} 
Our objective is to discover which object is being referred to in a $3$D space given a metric and semantic query.
Given a metric-semantic spatial instruction $q$ and a partial $3$D scene graph with objects as vertices and edges as relationships,
$\Gamma = (V,E)$, we infer a probability density $P(x \mid q, \Gamma)$ over 
$x \in \Omega_{\text{free}} \subset \mathbb{R}^3$ such that high-density regions 
correspond to locations satisfying the semantic referents, spatial predicates, 
and metric constraints specified in $q$. This density is then grounded to resolved 
object instances $(V)$ in the graph.
We introduce the Multi-Agent Probabilistic Grounding (MAPG) framework, a modular spatial reasoning system designed to ground compositional natural language queries in continuous $3$D environments.
MAPG comprises multiple interacting agents  (Fig. \ref{fig:method_overview-h}) that communicate over a shared memory ledger, that also retrieves the scene graph $\Gamma$, a referent belief state $B_t$, and the active spatial likelihood fields, derived from active constraints. This work draws inspiration from the (G$^3$) framework~\cite{tellex2011symbolgrounding}, and recent agent-based reasoning methods~\cite{han2024interpretinteractivepredicatelearning}. It has the following components:\\
\begin{figure}[t]
    \centering
    \vspace{2mm}
    \begin{subfigure}[t]{0.45\linewidth}
        \centering
        \includegraphics[width=0.95\linewidth]{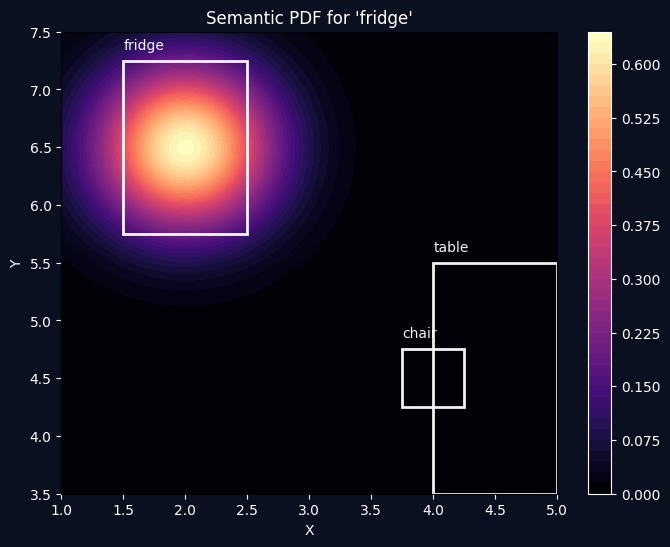}
        \caption{Semantic Grounding}
        \label{fig:fridge_semantic}
    \end{subfigure}\hfill
    \begin{subfigure}[t]{0.45\linewidth}
        \centering
        \includegraphics[width=0.95\linewidth]{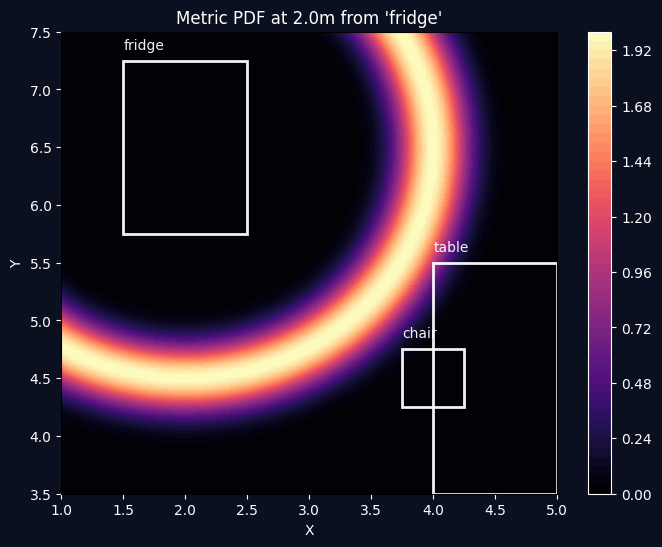}
        \caption{Metric Kernel}
        \label{fig:fridge_metric}
    \end{subfigure}

    \vspace{0.3em}

    \begin{subfigure}[t]{0.45\linewidth}
        \centering
        \includegraphics[width=0.95\linewidth]{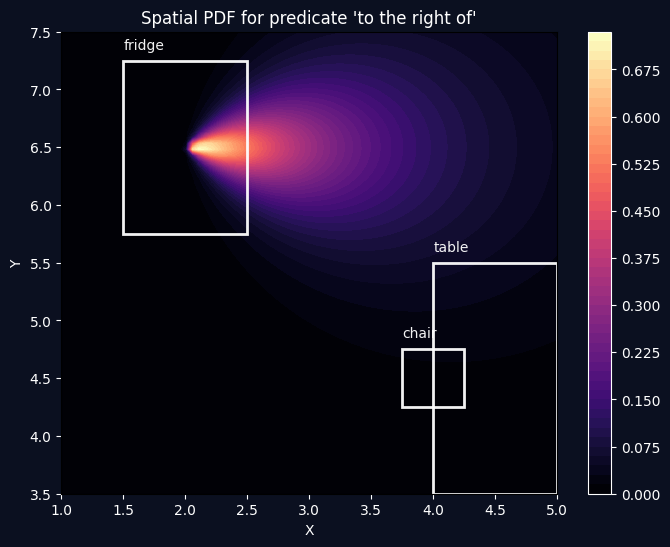}
        \caption{Spatial Kernel}
        \label{fig:fridge_spatial}
    \end{subfigure}\hfill
    \begin{subfigure}[t]{0.45\linewidth}
        \centering
        \includegraphics[width=0.95\linewidth]{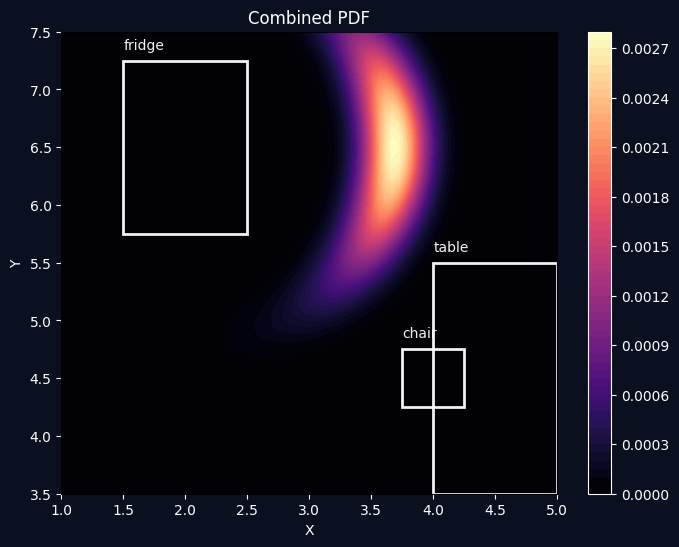}
        \caption{Composed Kernel}
        \label{fig:fridge_composed}
    \end{subfigure}

    \caption{MAPG grounding for the query “Where is \SI{2}{m} to the right of the fridge?” Semantic grounding identifies the fridge, the metric kernel models the \SI{2}{m} offset, and the directional spatial kernel captures the predicate “right of.” Their analytical composition yields a planner-ready goal distribution over feasible target locations.}
    \label{fig:fridge_kernels_row}
\end{figure}
\noindent$\!\!$1) \uline{\textbf{The Orchestrator:}} Responsible for parsing free-form natural language instructions into a sequence of symbolic clauses, called Spatial Description Clauses (SDCs)~\cite{kollar2010toward}, which bind spatial predicates to concrete referents in the environment.
Given the query ``Where is 2 meters to the right of the fridge?”, the Orchestrator extracts the following structured representation:
\textbf{\textit{Anchor object:}} \texttt{fridge}, \textbf{\textit{Spatial predicate:}} \texttt{right-of}, \textbf{\textit{Metric constraint:}} \texttt{2.0 meters}

In more complex instructions, multiple clauses may be extracted, forming a conjunction of spatial constraints that are composed downstream. The Orchestrator uses dependency parsing and task-specific templates to identify and align referents, predicates, and metrics.
These symbolic clauses are passed to the Grounding and Spatial agents for referent resolution and $3$D distribution generation, respectively.

\noindent$\!\!$2) \uline{\textbf{{Grounding Agent:}}} Responsible for resolving symbolic referents extracted by the Orchestrator (e.g., ``fridge'') into concrete object instances within the current environment. It interfaces directly with the semantic scene graph $\Gamma$, which represents the agent’s current world model.
Given a referent like ``the fridge,'' the agent queries all nodes in $\Gamma$ with matching or similar semantic labels using (i) string similarity between the referent text and each node’s stored label, (ii) CLIP-based similarity between the referent text and the candidate object's image crop \cite{radford2021learningtransferablevisualmodels}, and (iii) a spatial salience prior (such as proximity or visibility relative to the agent). 

A belief distribution $B_t(o)$ over referents is updated on the memory ledger, enhancing downstream reasoning. This grounding mechanism ensures that symbolic predicates produced by the Orchestrator are anchored to concrete, spatially localized objects in the scene.

\noindent3) \uline{\textbf{Spatial Agent}:}
Once a referent object $o_r$ is resolved, the Spatial agent generates a continuous probability density function (PDF) $P(x \mid r)$ over $3$D space, representing the likelihood of a goal location $x \in \mathbb{R}^3$ given a spatial predicate $r$ (e.g., ``right of").
Each metric semantic predicate describes a PDF over the physical space in our framework. We represent these PDFs as parametric kernels such as the von Mises or radial Gaussian kernels \cite{tellex2011symbolgrounding}, and we learn the parameters of these kernels using VLMs. We chose these kernels to represent the PDFs  because they functioned well to represent physical spaces in our experiments. However, this choice of kernels is a domain-specific choice.

Predicates depend not only on the world frame but also on object-centric frames. For example, the ``front'' of a chair is defined relative to its affordance, namely the direction in which a person would face when seated. Let $R_o \in \mathbb{R}^{3 \times 3}$ be the rotation matrix that maps directions from the object-local frame to the world frame, and $t_o \in \mathbb{R}^3$ be the anchor object position in world coordinates. The spatial field is therefore defined in the object's local frame and projected into the world frame. Each spatial predicate is modeled using a composition of two parametric kernel families. The spatial agent learns and specifies the parameters for these kernels. 
\\\textbf{Spatial kernel formulation:}
The directional kernel is defined as a von Mises–Fisher distribution: \\$P_{\text{dir}}(x)=\frac{1}{Z}\exp\!\left(\kappa\,\bigl(R_o\,m(\theta_0,\phi_0)\bigr)^\top \widehat{(x-t_o)}\right)$, where $\widehat{(x-t_o)}=\frac{x-t_o}{\|x-t_o\|}$ is the unit direction from the anchor to $x$, $Z$ is defined to normalize the distribution, and $\kappa$ controls directional concentration. The corresponding directional log-likelihood is $\ell_{\text{pred}}(x;\theta_0,\phi_0,\kappa)=\log P_{\text{dir}}(x)$.
\\\textbf{Metric kernel formulation:}
Given a resolved referent object $o_r$ with center $t_o \in \mathbb{R}^3$ and a spatial query of the form ``$d$ meters $r$ of $o_r$,'' the metric likelihood of a candidate point $x \in \mathbb{R}^3$ is modeled using a radial Gaussian: $\ell_{\text{met}}(x)=-\frac{1}{2\sigma_m^2}\left(\|x-t_o\|-d_0\right)^2$. Here, $d_0$ is the predicted mean offset and $\sigma_m$ controls tolerance to metric deviation.
The full combined log-likelihood across spatial components is then $\log P(x)=\ell_{\text{met}}(x;d_0,\sigma_m)+\ell_{\text{pred}}(x;\theta_0,\phi_0,\kappa)$.
An example of the kernels are shown in Fig. \ref{fig:fridge_kernels_row}.

\noindent4)~\uline{\textbf{Cascading Spatial Kernels:}} 
For a compound instruction like “Place the cup near the sink and to the left of the microwave,” two spatial kernels are constructed, one centered on the sink and another on the microwave. These fields are warped into global coordinates, added in log-space, and normalized to produce a multimodal density reflecting both predicates. This process is a close approximation to multiplying and normalizing the two PDFs to compose them, similar to prior compositional work \cite{du2024compositionalgenerativemodelingsingle}. If the two kernels align, their contributions reinforce; if they conflict, the resulting density distributes mass between satisfying regions.

\noindent5)~\uline{\textbf{Goal Selection and Planning Interface:}}
Once composed, \( P(x) \) serves as a goal likelihood map from which a planner can extract navigation targets via importance sampling or peak estimation. In practical implementations, we pass the top-$k$ sampled waypoints from \( P(x) \) to a sampling-based planner (RRT*) to generate executable trajectories.

The MAPG framework offers an interpretable, modular approach to spatial language grounding in embodied $3$D environments. It decomposes open world instructions, resolves referents generated by these instructions, generates spatial distributions for each of the referents, and analytically composes the predicates. MAPG provides both reasoning and planner-ready outputs for open world instructions. The following sections describe the experimental setup and benchmarks used to evaluate MAPG’s spatial capabilities.

\section{Datasets}
\label{sec:datasets}

Several benchmarks have been proposed to study spatial reasoning, including SpatialRGPT-Bench \cite{spatialRGPT} and $3$DSR-style \cite{3dsrbench} benchmarks. However, their primary modality is \emph{single images} (or image-centric VQA), which limits evaluation of (i) viewpoint consistency, (ii) partial observability, and (iii) the need for exploration before answering.
In parallel, temporal $3$D scene representations such as semantic scene graphs have been argued to provide persistent memory for embodied reasoning agents, enabling more robust grounding of object locations and relations over time.
While embodied QA benchmarks such as GraphEQA (HM-EQA)~\cite{graphEQA} move closer to this problem by emphasizing interaction and persistent memory, they lack \emph{metric--semantic--predicate} grounding queries typical of human interactions.

To explicitly evaluate $3$D grounded spatial reasoning, we introduce \textbf{MAPG-Bench} (Map-based Anchored Predicate Grounding Benchmark). This human-annotated dataset tests metric grounding, predicate interpretation, and anchor disambiguation in navigable indoor environments. It requires allocentric localization for ``Where'' queries, exposes viewpoint variation through embodied exploration, and evaluates compositionality via multi-anchor relations.

MAPG-Bench uses $30$ HM$3$D~\cite{ramakrishnan2021habitatmatterport3ddatasethm3d} indoor house scenes rendered in Habitat-Sim. The authors collected annotations using an interactive tool that overlays semantic instance labels on the agent's egocentric view. Annotators explored the environment, selected anchor instances, and recorded the agent's pose. For object-to-world questions, the ground truth answer is an explicit \emph{$3$D allocentric point} placed on a navigable surface. This represents an actionable navigation waypoint rather than a textual description.
Questions were generated semi-automatically to ensure diversity and feasibility. Predicates (left, right, near, between, etc.) were sampled from a fixed set, and metric distances were sampled at typical indoor scales ($0.5$ to $4.0$ meters).
We included a specialized subset of prompts to probe known VLM failures, such as resolving the ``front'' of an object versus the camera's ``front''. This subset was constructed to teach and test the meaning of such ambiguous predicates to future VLMs.
\begin{figure}[!t]
  \centering
  \vspace{2mm}
  \includegraphics[width=0.90\linewidth,height=10cm,keepaspectratio]{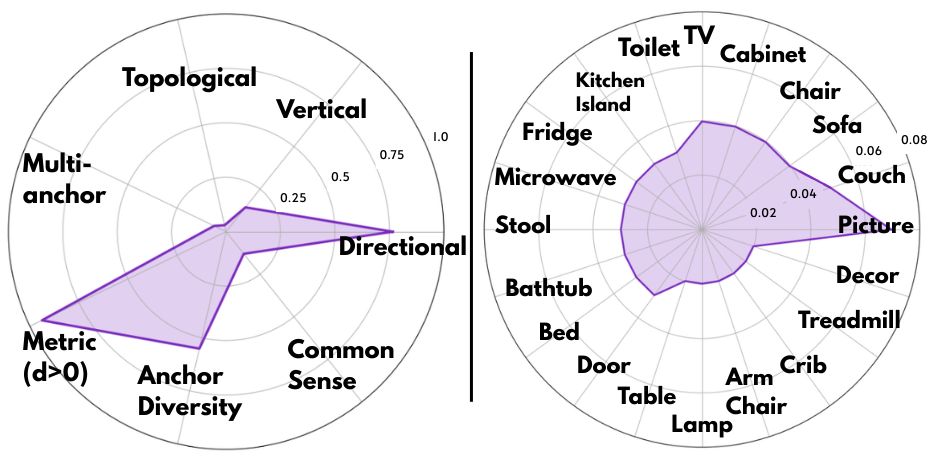}
  \caption{\textbf{Distribution} of query categories (left) and anchor object labels (right)  in MAPG-Bench.
  }
  \label{fig:mapg_distributions}
\end{figure}
\section{Experimental Results}
\label{sec:experiments}

\begin{table*}[t]
\vspace{4pt}
\centering
\small
\setlength{\tabcolsep}{5pt}
\renewcommand{\arraystretch}{1.15}

\resizebox{\linewidth}{!}{
\begin{tabular}{lccccccccc}
\toprule
\textbf{Method}
& \textbf{$d$ Err. O-O $\downarrow$} & \textbf{$d$ Err. O-W $\downarrow$}
& \textbf{Yaw Err. $\downarrow$} & \textbf{Pitch Err. $\downarrow$}
& \textbf{Obj. Sel. $\uparrow$} & \textbf{Common Sense $\uparrow$}
& \textbf{TSR $\uparrow$} & \textbf{Anchor Pick SR $\uparrow$} & \textbf{Avg. Traj. Len $\downarrow$} \\
& \textbf{(\si{\meter})} & \textbf{(\si{\meter})}
& \textbf{(\si{\degree})} & \textbf{(\si{\degree})}
& \textbf{(Acc.)} & \textbf{(Acc.)}
& \textbf{(\%)} & \textbf{(\%)} & \textbf{(\si{\meter})} \\
\midrule


\multicolumn{10}{l}{\textbf{Open-Sourced Specialist}} \\
\midrule
SRGPT        & $0.5$ & N/A & N/A & N/A & $0.36$ & N/A & N/A & N/A & N/A \\
GraphEQA     & N/A & $5.82$ & $13.5$ & $27.9$ & $0.34$ & 0.14 & 0.78 & 0.48 & 1.32 \\
\midrule

\multicolumn{10}{l}{\textbf{Ours}} \\
\midrule
MAPG (Gemini 3 Pro)        & $0.58$ & \textbf{$0.08$} & $13.8$ & $6.2$ & $0.30$ & 0.14 & $0.46$ & $0.46$ & $14.6$ \\

MAPG (OpenAI GPT 5.2) & $0.42$ & $\textbf{0.07}$ & $1.9$  & $4.4$ & $0.28$ & $0.16$ & $\textbf{0.98}$ & $\textbf{0.98}$ & $\textbf{1.3}$ \\

MAPG (Gemini 2.5 Pro)      & $0.08$ & $0.45$ & $4.9$  & $\textbf{3.8}$ & $\textbf{0.42}$ & $0.08$ & $0.9$ & $0.94$ & $2.5$ \\

\textbf{MAPG (Claude Opus 4.6)$^{*}$}     & $\textbf{0.07}$ & $0.43$ & $\textbf{0.3}$ & $10.3$ & $0.36$ & \textbf{0.30} & $\textbf{0.98}$ & $0.95$ & $1.8$ \\
\bottomrule
\end{tabular}
}

\caption{ Comparison of MAPG and base model performance on \textbf{MAPG-Bench}.
Proprietary generalist models were benchmarked; they exhibit inconsistent performance and do not reliably produce actionable $3$D goals across tasks, so we omitted them for space (available on our website).
$^{*}$ represents best overall model (highlighted among Methods).
Metrics with $\uparrow$ are higher-is-better; $\downarrow$ are lower-is-better.
$d$ Err.\ (O-O/O-W) measures goal localization precision relative to the anchor and in the global frame, while yaw/pitch errors measure directional consistency of the predicted offset (e.g., left/right/front/behind).
\textbf{Obj.\ Sel.} evaluates whether the predicted \emph{target} satisfies the query (i.e., the agent selects a goal consistent with the intended spatial relation), \textbf{Anchor Pick SR} measures anchor \emph{instance disambiguation} under partial observability.
\textbf{TSR} captures end-to-end reliability of completing the instruction, and \textbf{Avg.\ Traj.\ Len} measures behavioral efficiency.}
\label{tab:mapg_bench}
\end{table*}

We evaluate \emph{goal grounding} under a common requirement: given a spatial query, egocentric observations, and a persistent $3$D scene representation (scene graph), the system must output a grounded target that is directly usable by a planner. Depending on the query type, the output is either (i) an \emph{allocentric waypoint} for object-to-world (\emph{``Where''}) queries or (ii) an \emph{object selection} for object-to-object queries. Our evaluation is designed to isolate whether a method can (a) resolve the intended anchor under ambiguity, (b) interpret predicates in a consistent frame of reference, and (c) satisfy explicit metric constraints, while remaining robust under viewpoint changes and partial observability. 

We compare against state-of-the-art models for embodied question answering such as GraphEQA (Q\&A over a semantic graph)  and SRGPT (spatial specialist model) along with Standard VLM baselines. We modified GraphEQA's planner interface to answer \emph{spatial ``where'' questions} by producing a text waypoint (or equivalent allocentric target) from the scene graph, which yields a fair scene-graph-driven baseline for metric-semantic grounding. 


On MAPG-Bench we report both geometric error and task-level success. We measure distance error for object-to-object (O--O) and object-to-world (O--W) grounding (meters), and yaw/pitch error (degrees) to quantify directional consistency of the predicted offset. For identification-style cases, we report object-selection accuracy using embedding-based soft matching between predicted and ground-truth object names to account for label-space mismatches between HM$3$D annotations and Hydra-derived semantic classes; MAPG is scored using the best match over its ranked Top-1/Top-2 predictions, while baselines are scored using their single selected object. For embodied efficiency, we report task success rate (TSR), anchor-pick success rate (instance disambiguation under partial observability), and average trajectory length (meters) as a proxy for how much exploration is required before committing to a grounded target. We compare MAPG against three categories of baselines: open-sourced generalist VLMs, open-sourced spatial specialists, and proprietary VLMs. We evaluate multiple MAPG variants that share the same distributional grounding pipeline but differ only in the underlying foundation model used by the grounding components, isolating the effect of model choice from pipeline structure.

\noindent A) \textbf{MAPG-Bench results:}
Table~\ref{tab:mapg_bench} reports quantitative performance on MAPG-Bench. Relative to the scene-graph EQA baseline, MAPG achieves large reductions in object-to-world localization error and directional error, which directly supports our central claim: \emph{treating grounding as composition of metric + predicate likelihoods yields metrically consistent waypoints rather than nearest-satisfying heuristics.}

GraphEQA exhibits a large O--W distance error ($5.82\,m$) and high angular inconsistency ($13.5$$^\circ$ yaw, 27.9$^\circ$ pitch), reflecting its tendency to select the closest object/region satisfying a predicate without enforcing the metric offset. MAPG (OpenAI GPT-5.2) reduces O--W error from $5.82\,m$ to $0.07\,m$ ($98.8\%$ reduction) with an $SEM$ \tn{undefined} of $0.03m$, yaw error from 13.5$^\circ$ to 1.9$^\circ$ (85.9\% reduction), and pitch error from $27.9$$^\circ$ to $4.4$$^\circ$ ($84.2\% $ reduction), while achieving $0.98$ Task SR and $0.98$ anchor-pick SR with short trajectories ($1.3\,m$). These gains align with MAPG’s two design novelties: (i) explicit query decomposition (anchor / predicate / metric), and (ii) probabilistic composition of expert outputs into a planner-queryable goal distribution.

On object-to-object grounding, SRGPT achieves $0.50\,m$ error, consistent with its ability to reason about close-up relations in a single view. However, MAPG (Claude Opus 4.6) reaches $0.07\,m$ O--O error (an $86\%$ reduction vs $0.50\,m$), indicating that anchoring spatial kernels to resolved $3$D object instances yields more precise metric satisfaction beyond single-image heuristics.

We observe that model choice affects the operating point (e.g., MAPG variants trade off O--O vs O--W error), but the overall pattern persists: MAPG variants consistently achieve low metric error and improved directional consistency when compared to the scene-graph planner baseline, validating that the \emph{pipeline structure} (decomposition + composition) is the primary driver of spatial grounding performance.

To contextualize MAPG as a component in broader embodied question answering, we also benchmark on HM-EQA style QA where success is measured by multiple-choice answer accuracy rather than explicit waypoint grounding. This benchmark primarily tests the \emph{ anchor object selection} capability and semantic state inference, and is therefore complementary rather than a substitute for MAPG-Bench. 

\begin{table}[t]
\centering
\footnotesize
\setlength{\tabcolsep}{5pt}
\renewcommand{\arraystretch}{1.03}

\begin{tabular}{p{4.8cm}cc}
\toprule
\multicolumn{3}{c}{\textbf{Benchmarking on HM-EQA Question Answering}} \\
\midrule
\textbf{Method} & \textbf{TSR $\uparrow$} & \textbf{$L_{\tau}$ (\si{\m}) $\downarrow$} \\
\midrule

Explore-EQA & $0.52$ & $38.1$ \\
Explore-EQA-GPT4o & $0.46$ & $6.3$ \\
Explore-EQA-Llama4-Mav & $0.44$ & $10.4$ \\
Explore-EQA-Gemini-2.5Pro & $0.54$ & $12.3$ \\
GraphEQA-Gemini-2.5Pro & $\textbf{0.67}$ & $7.41$ \\
GraphEQA (Gemini 2.5 Pro)$^{+}$ & $0.61$ & $5.48$ \\
GraphEQA (GPT 5.2)$^{+}$ & $0.63$ & N/A \\
MAPG (OpenAI GPT 5.2) (Ours) & $0.60$ & N/A \\
MAPG (Claude 4.6 Opus) (Ours) & $\textbf{0.71}$ & $6.62$ \\
\bottomrule
\end{tabular}
\caption{EQA question-answering accuracy comparison. $^{+}$ indicates our implementation of the particular baseline. While MAPG achieves higher TSR, we also have longer Trajectory length, consistent with our system, which carries out more explorations before deciding on the target object.}
\label{tab:eqa_qa_accuracy}
\end{table}

Table~\ref{tab:eqa_qa_accuracy} reports QA accuracy. GraphEQA achieves 0.67 accuracy, MAPG variants are competitive. This is an indication of anchor grounding module performs comparably on localizing the anchor objects and evidence aggregation are sufficiently strong to transfer to EQA-style answering. However, HM-EQA rewards semantic state inference and dataset-specific QA priors rather than metrically grounded waypoint prediction. We therefore use HM-EQA as a complementary diagnostic for the object-selection submodule, while MAPG-Bench remains our primary evaluation for metric-semantic goal grounding in continuous $3$D space.

\noindent B) \textbf{Ablations:}
We ablate MAPG to quantify why explicit spatial reasoning and modular agents are necessary. Table~\ref{tab:ablations_spatial} shows that removing the explicit spatial reasoner (replacing it with a single ``fat'' planner with curated chain-of-thought) drops object selection SR from $0.42$ to $0.20$, while the full MAPG configuration improves over the GraphEQA base ($0.34$ to $0.42$). These results support the claim that \emph{explicit composition of metric and predicate constraints} is not recoverable from prompting alone.

We further evaluate robustness under uncertainty induced by occlusion. In Table~\ref{tab:ablations_occlusion}, the explicit spatial reasoner improves object selection SR from $0.30$ to $0.50$ under occlusion, indicating that maintaining intermediate belief over anchors and deferring commitment can recover from partial observability when the anchor eventually enters the map.

Finally, we include a commonsense-focused subset (Table~\ref{tab:mapg_bench}) to stress frame-of-reference and pragmatic failures. Performance here highlights a limitation: commonsense QA-style inference is not the target strength of MAPG’s current design, reinforcing that MAPG’s value is in producing \emph{metrically consistent, planner-ready targets} for spatial queries rather than solving general semantic QA.

\noindent C) \textbf{Robot Demo:}
To demonstrate real-world feasibility, we constructed a scene graph from a physical indoor environment and run inference with the MAPG on a set of three spatial queries. MAPG successfully grounded the queried targets in the real-world environment, indicating that the method transfers beyond simulation when a structured scene representation is available. However, scaling this evaluation to a systematic benchmark would require collecting a larger real-world dataset with consistent scene graph annotations and ground-truth spatial labels. A video of the robot demonstration is provided in the supplementary material.

\begin{table}[t]
\centering

\renewcommand{\arraystretch}{1.1}
\setlength{\tabcolsep}{6pt}

\begin{tabular}{l c}
\toprule
\textbf{Method} & \textbf{Object selection SR $\uparrow$} \\
\midrule
GraphEQA (base)                          & $0.34$ \\
MAPG (CoT w/o spatial reasoner)            & $0.20$ \\
MAPG (explicit spatial reasoner)         & $0.42$ \\
\bottomrule
\end{tabular}

\caption{Replacing the explicit spatial reasoner with CoT degrades object selection success rate, while the full MAPG variant achieves the best performance. }
\label{tab:ablations_spatial}
\end{table}

\begin{table}[t]
\centering

\renewcommand{\arraystretch}{1.1}
\setlength{\tabcolsep}{6pt}

\begin{tabular}{l c}
\toprule
\textbf{Method} & \textbf{Object selection SR $\uparrow$} \\
\midrule
GraphEQA (base)                          & $0.30$ \\
MAPG (CoT- w/o spatial reasoner)            & $0.30$ \\
MAPG (explicit spatial reasoner)         & $0.50$ \\
\bottomrule
\end{tabular}

\caption{Occlusion highlights the limits of CoT-only reasoning, while the full MAPG variant performs best.}
\label{tab:ablations_occlusion}
\end{table}




\begin{figure*}[!t]
  \centering
  \includegraphics[width=\textwidth]{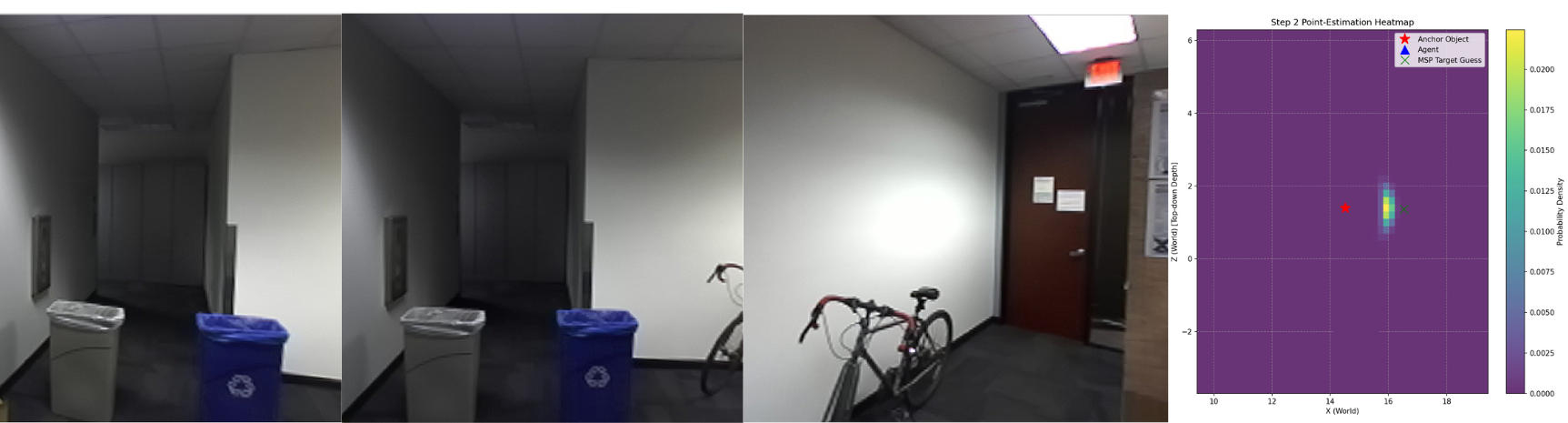}
  \caption{\textbf{Real-world MAPG grounding example.} Using observations collected from a Robotis AI Worker robot, we construct an offline scene graph of a physical indoor environment. Three of those image frames are demonstrated here, with the query “Where is $1$ meter to the right of the trash can close to the bicycle?” MAPG infers a spatial probability distribution over candidate goal locations (right). The anchor object (red) and predicted goal location (green) illustrate metrically consistent semantic grounding beyond simulation. The grounded area was $1$ meter to the right of the blue trash can, which is also the general region of the bicycle.} 
  \label{fig:obs-anchor-trace}
\end{figure*}

\section{Observations and Discussions}
\label{sec:observations}

MAPG’s improvements are best explained by two design choices: \textbf{(i) query decomposition} (anchor / predicate / metric) and \textbf{(ii) probabilistic composition} of expert outputs into a planner-queryable goal density. On MAPG-Bench (Table~\ref{tab:mapg_bench}), these choices translate into large gains over the strongest open-sourced scene-graph baseline (GraphEQA): MAPG (GPT-5.2) reduces O--W error from $5.82\,m$ to $0.07\,m$  and improves Task SR from $0.78$ to $0.98$, while keeping trajectories short ($1.32\,m$ $\rightarrow$ $1.3\,m$). Directional consistency improves in the same regime with a drop in yaw and pitch error. Ablations indicate these gains are primarily structural rather than “better prompting” where removing the explicit spatial reasoner reduces object selection SR from $0.42$ to $0.20$ (Table~\ref{tab:ablations_spatial}), and under occlusion, explicit reasoning improves object selection SR from $0.30$ to $0.50$ (Table~\ref{tab:ablations_occlusion}).



\noindent1){\textbf{ Multi-view anchor disambiguation acts as evidence accumulation:}}
Fig.~\ref{fig:obs-anchor-trace} shows MAPG deferring anchor commitment until evidence is sufficient (e.g., searching for the red kettle across views). Empirically, this behavior aligns with MAPG’s high Anchor Pick SR (up to $0.98$) and high Task SR ($0.98$) with minimal additional exploration ($1.3\,m$ average trajectory). In contrast, GraphEQA’s pipeline often commits immediately to the nearest plausible proxy under partial observability, which is consistent with its lower Task SR ($0.78$). The occlusion ablation strengthens this interpretation: explicit spatial reasoning increases object-selection SR from $0.30$ to $0.50$, suggesting that maintaining intermediate belief and re-checking hypotheses is what recovers performance when the anchor is not reliably visible.

\noindent2){\textbf{ Composed belief maps produce planner-ready waypoints:}} MAPG produces \emph{planner-ready outputs} because the final target is derived from a composed spatial belief over free space rather than a single textual guess. The anchor location, metric constraint, and predicate direction are each represented as structured likelihoods and composed into a continuous density over $\Omega_{\text{free}}$. A waypoint is then extracted via peak estimation or top-$k$ sampling from this density, yielding an allocentric coordinate that directly satisfies geometric feasibility and can be consumed by a motion planner without additional interpretation. This explains the dominant quantitative gap with GraphEQA: GraphEQA’s large O--W error ($5.82\,m$) indicates a nearest-satisfying bias where direction may be satisfied but distance is not enforced, whereas MAPG’s O--W errors are near-zero for the best variants (e.g., $0.07\,m$ for GPT-5.2; $0.08\,m$ for Gemini 3 Pro). Likewise, the sharp reduction in yaw/pitch errors (to $1.9^\circ/4.4^\circ$ for GPT-5.2) is consistent with representing predicate effects as an explicit directional term in a shared allocentric frame, rather than as an implicit language-only guess.

\noindent3){\textbf{ Where MAPG helps most:}}
MAPG’s gains concentrate on queries requiring \textbf{joint metric + relational satisfaction} under viewpoint variation. For object-to-object grounding, MAPG (CO 4.6) achieves $0.07\,m$ O--O error versus SRGPT’s $0.50\,m$, indicating that anchoring metric kernels to resolved $3$D instances yields more precise grounding than single-image heuristics. Across MAPG variants there is a stable overall pattern: low metric error, improved angular consistency, and high Task SR relative to the scene-graph planner baseline, supporting our claim that \emph{decomposition + composition} is the primary driver.

\noindent4){\textbf{ HM-EQA: competitive performance without optimizing for MCQ priors:}}
HM-EQA complements MAPG-Bench by testing anchor/object selection in a multiple-choice QA regime. Table~\ref{tab:eqa_qa_accuracy} shows GraphEQA-Gemini-2.5Pro at $0.67$ and image-only Explore-EQA baselines at $0.44$--$0.54$. MAPG remains competitive: MAPG (OpenAI GPT-5.2) achieves $0.60$, and MAPG (Claude 4.6 Opus) reaches $0.71$ (\textbf{+0.04} over $0.67$). We view this as a transfer signal that MAPG’s verification/evidence mechanisms do not harm EQA competency, while MAPG-Bench remains the benchmark that isolates MAPG’s primary contribution: \emph{planner-ready metric-semantic ``where'' grounding}.

\noindent5) {\textbf{Failure modes and limitations}}
Most MAPG failures stem from limitations in the underlying scene representation rather than the grounding pipeline. The dominant issue is \emph{scene-graph incompleteness}, where occluded objects never enter the map, forcing the system to ground relative to an incomplete world model. Errors also arise from \emph{frame-of-reference ambiguity}, particularly for predicates such as ``in front of'' or ``behind'' when intrinsic object orientation is unclear. Despite this issue MAPG substantially reduces angular inconsistencies. Finally, misalignment between the free-space map and scene graph can blur the composed belief map and reduce waypoint precision. These limitations highlight the importance of accurate mapping and representation for reliable spatial grounding.

\section{Conclusion}
In this work, we presented MAPG, a probabilistic grounding framework for translating metric-semantic language into planner-ready goals in continuous $3$D Environments. Rather than treating grounding as a single hard decision, our approach decomposes a query into anchor, predicate and metric components, grounds them against an online $3$D Scene Graph and egocentric observations, and composes their outputs into a continuous goal distribution. We also introduced MAPG-Bench, a benchmark designed specifically for metric-semantic goal grounding in embodied settings, covering $30$ HM$3$D Scenes and $100$ annotated queries. We demonstrate that across this benchmark, MAPG has consistently produced more metrically accurate and spatially consistent outputs than other spatial reasoning and scene-graph based baselines \cite{spatialRGPT, graphEQA}, reducing the object-to-world grounding error from \SI{5.82}{m} to \SI{0.07}{m} and improving task success rate from $0.78$ to $0.98$ in our best setting. Our ablations further show that these gains come from methodological gains such as explicit decomposition and probabilistic composition, not prompting alone, and our real-world zero-shot demo suggests that the approach can transfer beyond simulation. At the same time, failures arise from incomplete scene graphs, frame-of-reference ambiguity, and map alignment errors. 
Overall, this work argues that a distributional and compositional grounding approach provides a reliable interface between language understanding, spatial memory, and execution for open-world metric-semantic navigation.

\bibliographystyle{IEEEtran}
\bibliography{references}
\end{document}